# Interpretable Brain Age Prediction Using White Matter Fiber Orientation Distributions with Rotation-Equivariant Deep Learning

Chenyang Fan, Robert D. Sanders, Elie Matar, Yixuan Yuan, Xiang Li, Sharon L. Naismith, Fernando Calamante, and Jinglei Lv.

***Abstract*—Diffusion MRI (dMRI) provides information about white matter microstructure that can support brain age prediction. However, many existing prediction models use diffusion tensor imaging (DTI) scalar metrics that omit orientation information. Deep learning is increasingly used for brain age prediction, but conventional CNNs do not explicitly respect the geometric structure of orientation-dependent features. We present an interpretable brain age prediction framework that combines white matter fiber orientation distributions (WM-FODs) with an SE(3) (Special Euclidean group in 3 dimensions) equivariant CNN. The network operates directly on spherical harmonic (SH) coefficients of the FOD, preserving their transformation properties under rotation. Evaluation in 692 participants aged 36–90 years from the Human Connectome Project Aging dataset showed that the combined framework outperformed the evaluated DTI-based models and conventional CNNs. Averaging predictions from five cross-validation models yielded a mean absolute error of 3.54 years and Pearson's correlation of 0.958. Our results indicated that the performance advantage depended on combining WM-FODs with equivariant processing. Prediction accuracy was highest at a maximum SH order of Lmax = 2 and declined at higher orders. Tract-level occlusion, where each segmented tract was masked in turn, revealed age-dependent patterns of model sensitivity, with importance peaks in frontal projection pathways at 36–44 years, the fornix at 45–54 years, the callosal isthmus at 55–64 years, the middle cerebellar peduncle at 65–74 years, and the callosal genu at 75–90 years. These findings support rotation-equivariant processing of WM-FODs for accurate brain age prediction with anatomically interpretable model sensitivity.**

## I. INTRODUCTION

Brain age, estimated from neuroimaging, provides a quantitative measure of age-related brain characteristics and has emerged as a potential biomarker of brain health [1]. The difference between predicted brain age and chronological age, termed the brain age gap (BAG), has been associated with mortality risk [2], cognitive decline [3], and neurological disorders [4]. Early approaches primarily used T1-weighted structural MRI to characterize macroscopic anatomical changes, including atrophy. Diffusion MRI (dMRI) provides complementary information about white matter microstructure and the fiber pathways that support cognition.

Deep learning approaches to dMRI-based brain age prediction face two key limitations. First, many studies use scalar metrics derived from diffusion tensor imaging (DTI), such as fractional anisotropy (FA) [5], [6], [7]. Although convenient, these metrics omit orientation information in the diffusion signal, limiting their representation of complex fiber configurations [8]. Second, conventional convolutional neural networks (CNNs) do not explicitly account for the transformation properties of orientation-dependent dMRI features.

White matter fiber orientation distributions (WM-FODs) describe the distribution of fiber orientations within each voxel. They can be represented by spherical harmonic (SH) coefficients that transform together under rotation (Fig. 1). Within each SH order, rotating an FOD mixes the coefficients according to a prescribed transformation; the coefficients therefore have a shared geometric structure. Standard CNNs treat them as ordinary input channels without enforcing this structure. SE(3)-equivariant CNNs instead constrain their operations so that the resulting features transform consistently

C. Fan is with the School of Biomedical Engineering and Brain and Mind Centre at the University of Sydney, Sydney, NSW 2050 Australia (e-mail: cfan9905@uni.sydney.edu.au). R. D. Sanders is with the Brain and Mind Centre and Central Clinical School at the University of Sydney, Sydney, NSW, 2050 Australia (e-mail: robert.sanders@sydney.edu.au). E. Matar is with the Brain and Mind Centre and Central Clinical School at the University of Sydney, Sydney, NSW, 2050 Australia (e-mail: elie.matar@sydney.edu.au). Y. Yuan is with the Department of Electronic Engineering, Chinese University of Hong Kong, Hong Kong SAR 999077, China (e-mail: yxyuan@ee.cuhk.edu.hk). X. Li is with the Department of Radiology, Massachusetts General Hospital and Harvard Medical School, Boston, MA 02114, USA (e-mail: xli60@mgh.harvard.edu). S. L. Naismith is with the Brain and Mind Centre and School of Psychology at the University of Sydney, Sydney, NSW, 2050 Australia (e-mail: sharon.naismith@sydney.edu.au). F. Calamante is with the School of Biomedical Engineering, Brain and Mind Centre, and Sydney Imaging, at the University of Sydney, Sydney, NSW 2050 Australia (e-mail: fernando.calamante@sydney.edu.au). Corresponding author, J. Lv, is with the Sydney Medical School at the University of Sydney, Sydney, NSW 2050 Australia and School of Computer, Data and Mathematical Sciences, Western Sydney University, Parramatta, NSW 2116 Australia. (e-mail: jinglei.lv@sydney.edu.au).

with rotations of the input [9].

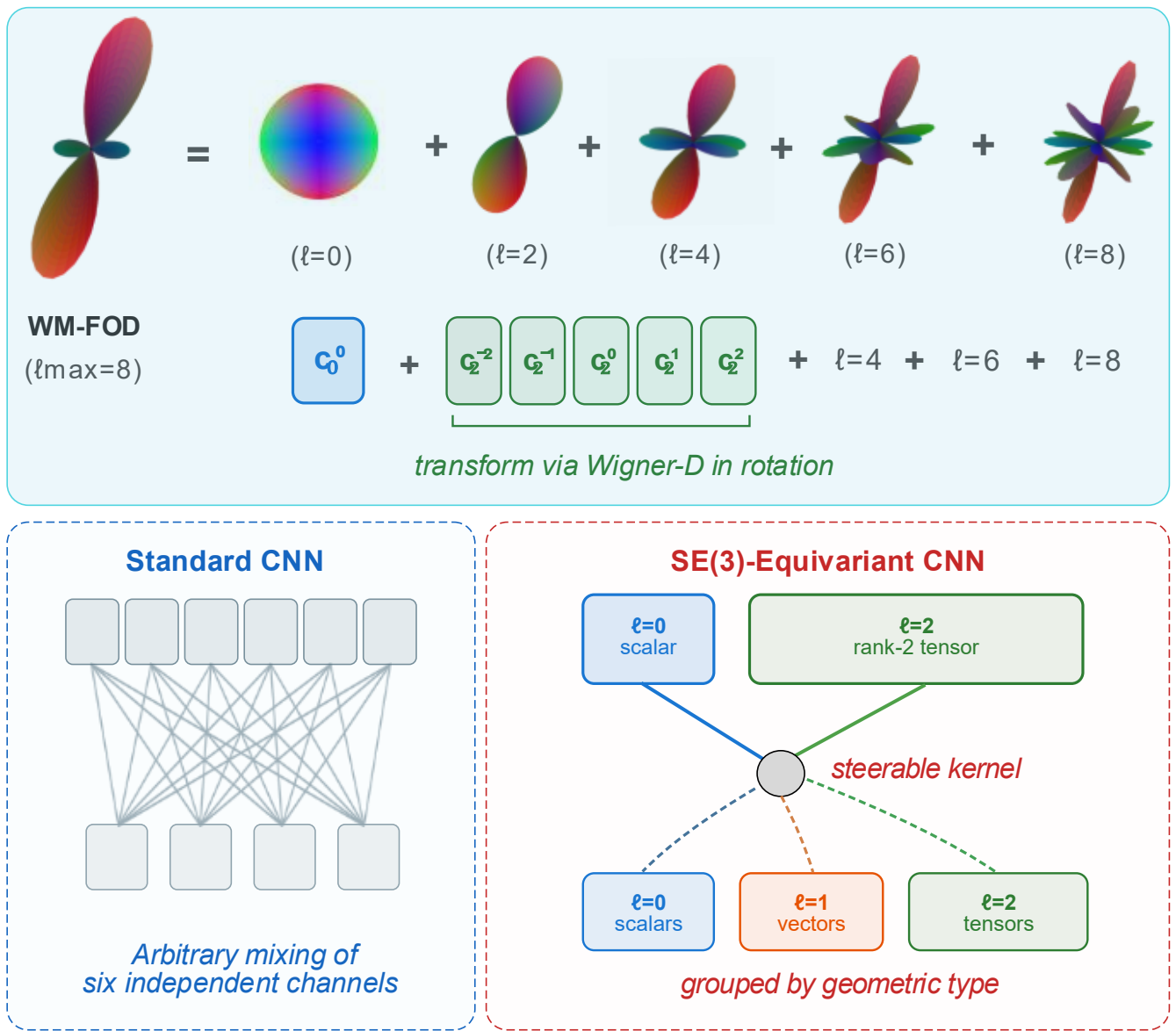

Fig. 1. WM-FOD representation and SE(3)-equivariant processing. A WM-FOD with Lmax = 8 is decomposed into even-order spherical harmonic components (ℓ = 0, 2, 4, 6, 8), which transform under rotation according to Wigner D-matrices. The SE(3)-equivariant CNN treats each order as a geometric field (e.g., ℓ = 0 for scalars and ℓ = 2 for second-order components) and combines these fields using steerable kernels. The resulting scalar, vector, and higher-order features retain their prescribed transformation properties.

Interpretability is also important for understanding the anatomical information used in brain age prediction. Fixed-size occlusion windows and voxel-wise saliency maps can be difficult to relate to individual white matter pathways. Anatomically informed occlusion offers a complementary approach: masking tract-defined regions allows the sensitivity of a trained model to be assessed in an anatomical context. Such analyses identify regions relevant to prediction, although they do not establish causal roles in aging.

This study makes three contributions:

1) Feature representation: We compare WM-FODs with conventional DTI-derived features and evaluate the effect of maximum SH order. WM-FODs combined with equivariant processing outperform the evaluated DTI-based models, with prediction accuracy peaking at Lmax = 2.

2) Architecture: We developed an SE(3)-equivariant CNN [9] that operates directly on WM-FOD SH coefficients. We evaluate its predictive accuracy, stability under input rotations, and computational cost against conventional CNNs in whole-brain analyses.

3) Interpretability: We use tract-level occlusion to quantify model sensitivity to anatomically defined white matter regions. Tract-level and volume-normalized importance measures identify pathways relevant to prediction and characterize how their importance varies across adult age groups.

## II. Related Work

### A. Brain Age Prediction Using Diffusion MRI

Previous dMRI-based brain age models in healthy aging cohorts have reported mean absolute errors (MAEs) of approximately 3.5–7.6 years, in samples ranging from 89 [10] to more than 30,000 [11] participants. Methods include linear and regularized regression [12], [13], [11], [14], [15], support vector machines [10], tree-based ensembles [5], [7], [16], [17], and deep learning [6], [18], [19]. Inputs have included DTI scalars, free-water DTI measures, advanced diffusion metrics, tract-wise summaries, and white matter connectome features. These approaches generally use scalar or handcrafted representations that do not preserve the full orientation structure of the diffusion signal.

Differences in cohort age range complicate comparisons across studies. A narrower age range can reduce MAE while also reducing the variance available to explain, resulting in a lower correlation between predicted and chronological age. For example, the lowest MAE among the studies compared here was 3.51 years over a 23-year age span, with Pearson's correlation (r) = 0.49 [13]. Studies spanning a broader adult age range generally reported higher MAEs but stronger correlations. Reporting both MAE and r, together with the cohort age range, therefore, supports more informative comparisons.

### B. From Scalar Metrics to WM-FODs

DTI and WM-FODs differ in both their orientation information and their modeling assumptions. DTI models diffusion within each voxel using a Gaussian tensor model and is susceptible to partial-volume effects, including contamination by cerebrospinal fluid (CSF). In contrast, multi-shell multi-tissue constrained spherical deconvolution (MSMT-CSD) [20] estimates WM-FODs that can represent complex fiber configurations while separating white matter, gray matter, and CSF signal components.

FOD-derived measures are widely used in fixel-based analyses to characterize fiber-specific white matter properties [21]. Their use as direct inputs to deep learning models for brain age prediction, however, remains relatively unexplored.

### C. Geometric Deep Learning for Diffusion MRI

Geometric deep learning incorporates known symmetries into neural network architectures. SE(3)-equivariant networks use steerable filters to ensure that features transform consistently under three-dimensional rotations [9]. For dMRI, Müller et al. introduced joint SE(3) × SO(3) convolutions on raw q-space signals [22], while Elaldi et al. developed E(3) × SO(3)-equivariant hemispherical convolutions for spherical deconvolution [23]. These methods demonstrate the value of equivariant processing for diffusion signal modeling and FOD estimation. Their focus differs from downstream prediction using precomputed WM-FOD representations.

In our framework, WM-FOD SH coefficient fields already encode angular information. A volumetric SE(3)-equivariant CNN can therefore process these fields directly, without the additional angular (SO(3)) convolution required by the joint spatial–angular methods above. We evaluate this approach for brain age prediction.

### D. Tract-Level Occlusion in Diffusion MRI Brain-Age Models

Regional and network-level occlusion analyses have identified anatomically interpretable sensitivities in structural MRI-based brain age models [24], [25], [26]. These studies have primarily examined gray matter or atlas-defined regions rather than diffusion-derived white matter pathways.

Previous dMRI-based brain age studies have used segmented bundles, tract groups, or along-tract diffusion summaries to estimate age and characterize bundle-specific aging patterns [14], [17]. Although these methods provide anatomical localization through predefined tract features, they do not directly test the sensitivity of a trained whole-brain model to tract-defined regions. We therefore apply tract-level occlusion as an anatomically informed, post-hoc analysis.

## III. Methods

Figure 2 summarizes the processing, prediction, and interpretation pipelines.

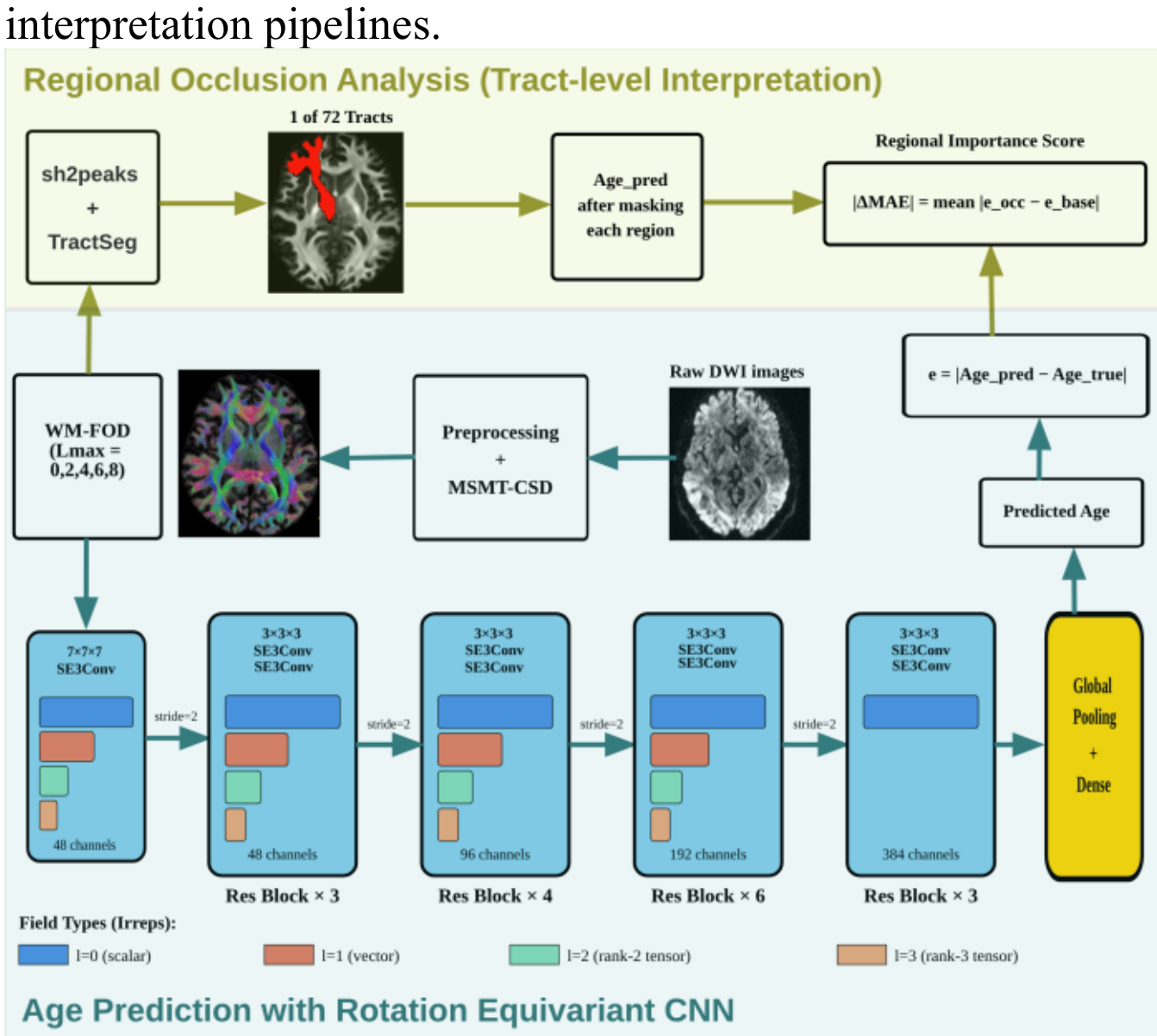

Fig. 2. Pipeline for brain age prediction and tract-level interpretation. Diffusion-weighted images are preprocessed and analyzed using MSMT-CSD to estimate WM-FODs at different maximum SH orders. SE(3)-ResNet34 processes the WM-FODs using equivariant convolutions, progressive channel expansion, and spatial downsampling, followed by global pooling and age prediction. Occlusion of 72 tract-defined regions yields importance scores (|ΔMAE|) that characterize the model's anatomical sensitivity.

### A. Dataset and Acquisition

Diffusion MRI data were obtained from Lifespan Release 2.0 of the Human Connectome Project-Aging (HCP-A) [27], which includes 725 participants. Participants older than 100 years (n=12) and those whose dMRI data failed preprocessing because of considerable residual imaging artifacts (n=21) were excluded. The final sample comprised 692 healthy participants aged 36–90 years (mean ± SD, 59.6 ± 14.9 years; 56% female). Data were acquired on a Siemens 3T Prisma scanner using a multi-shell protocol with 93 diffusion directions at b = 1500 s/mm² and 92 directions at b = 3000 s/mm², at 1.5-mm isotropic resolution. Alternating anterior-to-posterior and posterior-to-anterior phase encoding enabled susceptibility-distortion correction.

### B. Preprocessing

Preprocessing followed the MRtrix3 pipeline [28] and included denoising, Gibbs ringing removal, susceptibility-distortion and motion correction using FSL topup and eddy, and bias-field correction using ANTs N4. Images were upsampled to 0.8-mm isotropic resolution.

### C. Diffusion Feature Extraction

DTI metrics: The diffusion tensor was fitted using weighted linear least squares [29]. This yielded fractional anisotropy (FA), mean diffusivity (MD), axial diffusivity (AD), radial diffusivity (RD), and the six independent tensor elements. For early fusion, the four scalar maps were concatenated as input channels. For late fusion, a separate model was trained for each scalar metric, and the resulting age predictions were averaged at inference. Separately, the six tensor elements were evaluated as a six-channel input.

WM-FOD images: MSMT-CSD [20] was performed using the average of the tissue-specific response functions estimated for each participant with the Dhollander algorithm [30]. WM-FODs were computed at maximum SH orders of Lmax = 0, 2, 4, 6, and 8, corresponding to 1, 6, 15, 28, and 45 coefficient channels, respectively, to assess the optimal performance for age prediction.

### D. Model Architectures

*SE(3)-ResNet34*: We developed an SE(3)-equivariant convolutional framework based on the work of Weiler et al. [9] to construct a ResNet-34 architecture (Fig. 2). Input SH coefficients were mapped to their corresponding irreducible representations, with each even SH order ($\ell$ = 0, 2, 4, 6, 8) represented as a separate geometric field. Analogous to the channels of a conventional CNN, each hidden layer contained multiple feature fields of orders $\ell$ = 0–3 (Fig. 2), with fewer fields at higher orders. The maximum hidden order ($\ell \leq 3$) was chosen because it gave higher accuracy than $\ell \leq 1$ and $\ell \leq 5$ in preliminary experiments. Spatial downsampling used Gaussian low-pass filtering followed by strided subsampling. Group normalization, gated nonlinearities, and equivariant dropout (p = 0.01) were applied between layers. The final block retained only scalar ($\ell$ = 0) fields, which are invariant to rotation. Global mean pooling and a fully connected layer produced the final age estimate.

*Baseline CNNs*: We evaluated ResNet50 and ResNet152, adapted for three-dimensional volumetric inputs, and SFCN [31], a lightweight architecture designed for brain age prediction.

### E. Training Procedure

Participants were assigned to training (n = 588; 85%) and held-out test (n = 104; 15%) sets using stratified sampling to preserve the age distribution. Five-fold cross-validation was performed within the training set. Models were optimized to minimize MAE using Adam, with an initial learning rate of $1 \times 10^{-3}$ and weight decay of $1 \times 10^{-4}$. The learning rate was halved after 5 epochs without improvement in validation loss, down to a minimum of $1 \times 10^{-6}$. Early stopping used a patience of 15 epochs. Training used mixed precision on four Tesla V100 GPUs, with a per-GPU batch size of 4 for standard CNNs and 1 for SE(3)-ResNet34.

### F. Evaluation of Prediction Stability Under Rotation

To assess prediction stability under input rotations, we rotated the spatial WM-FOD volumes from −15° to +15° in 1° increments in the axial, coronal, and sagittal planes. SH coefficients were reoriented simultaneously using the corresponding Wigner D-matrices [32]. An illustrative test participant was selected for whom SE(3)-ResNet34 and ResNet50 had the same unrotated absolute prediction error (0.5 years). Stability was quantified as the standard deviation of predicted ages across rotations. We also evaluated a ResNet50 model trained with rotation augmentation (random rotations of ±1°, applied with probability 0.5) to compare architectural equivariance with augmentation-based robustness. All test participants (n = 104) were then evaluated at discrete rotations of 0°, ±5°, ±10°, and ±15° about each anatomical axis. Paired t-tests compared rotation-induced prediction variability between models.

### G. Regional Occlusion Analysis

Tract segmentation: TractSeg [33] was used to segment 72 anatomically defined white matter bundles, grouped as association, projection, commissural, cerebellar, or limbic pathways.

Occlusion procedure: For each tract, all voxels within its mask were set to zero, and SE(3)-ResNet34 predictions were recomputed across the full cohort (n = 692). Predictions for participants in the cross-validation sample (n = 588) were obtained from their respective out-of-fold models, so that no participant was evaluated by a model trained on that participant. For the held-out test set (n = 104), predictions were averaged across the five cross-validation models. Predicted ages were corrected for age-related bias using linear regression following Smith et al. [34] before calculating baseline and occluded MAE. The importance score was defined as $|\Delta MAE| = |MAE_{\{occluded\}} - MAE_{\{baseline\}}|$.

Per-voxel importance: To account for differences in tract size, |ΔMAE| was divided by tract volume (number of voxels); values are reported in units of $10^{-4}$ years/voxel.

Age-stratified analysis: Participants were grouped into five age bins: 36–44, 45–54, 55–64, 65–74, and 75–90 years (n = 143, 154, 137, 120, and 138, respectively). For each tract, mean absolute importance was calculated within each bin, and the bin with the highest importance was identified.

### H. Statistical Analysis

Prediction performance was evaluated on the held-out test set (n = 104) using MAE and Pearson's r. Single-model results are reported as mean ± SD across the five cross-validation models; ensemble results were obtained by averaging the five models' predictions for each participant. Models were compared using two-sided paired t-tests on participant-wise absolute prediction errors, and rotation stability was compared using paired t-tests on participant-wise SDs of predicted age across rotations. P-values were adjusted for multiple comparisons within each family of related tests using the Benjamini–Hochberg false discovery rate procedure ($q < 0.05$).

## IV. RESULTS

### A. Comparison of Brain Age Prediction Performance

#### 1) DTI-Derived Baseline Features

We first evaluated DTI-derived features using ResNet50 (Table 1). FA yielded the lowest MAE among the individual scalar metrics (5.36 ± 0.44 years). Using the six independent tensor elements (D_xx, D_xy, D_xz, D_yy, D_yz, D_zz) as input channels performed similarly to FA (5.48 ± 0.39 years; p = 0.70). Early fusion of all four scalar maps produced the best DTI-based performance (MAE = 4.97 ± 0.16 years; r = 0.916) and was used as the primary DTI benchmark.

TABLE 1. BRAIN AGE PREDICTION PERFORMANCE OF DTI-DERIVED FEATURES USING RESNET50

| DTI Feature | Test MAE (years) | Pearson's r |
|---|---|---|
| FA | 5.36 ± 0.44 | 0.8926 ± 0.0191 |
| MD | 5.83 ± 0.33 | 0.8830 ± 0.0132 |
| AD | 6.02 ± 0.28 | 0.8730 ± 0.0067 |
| RD | 6.32 ± 0.43 | 0.8696 ± 0.0131 |
| DTI tensor elements | 5.48 ± 0.39 | 0.8984 ± 0.0146 |
| Early fusion | 4.97 ± 0.16 | 0.9155 ± 0.0036 |
| Late fusion | 5.14 ± 0.23 | 0.9067 ± 0.0091 |

*Values are mean ± SD across the five cross-validation models on the held-out test set (n = 104).

#### 2) Effect of Maximum SH Order

We evaluated WM-FODs at five maximum SH orders (Lmax = 0, 2, 4, 6, 8) with SE(3)-ResNet34, ResNet50, ResNet152, and SFCN (Fig. 3; Table 2). All architectures showed a non-monotonic relationship between maximum SH order and accuracy, with the lowest prediction error at Lmax = 2; for ResNet50, for example, MAE increased from 4.66 years at Lmax = 2 to 5.57 years at Lmax = 8. Within SE(3)-ResNet34, Lmax = 2 performed significantly better than Lmax = 0, 6, and 8 ($p < 0.05$), but did not differ significantly from Lmax = 4 (p = 0.27).

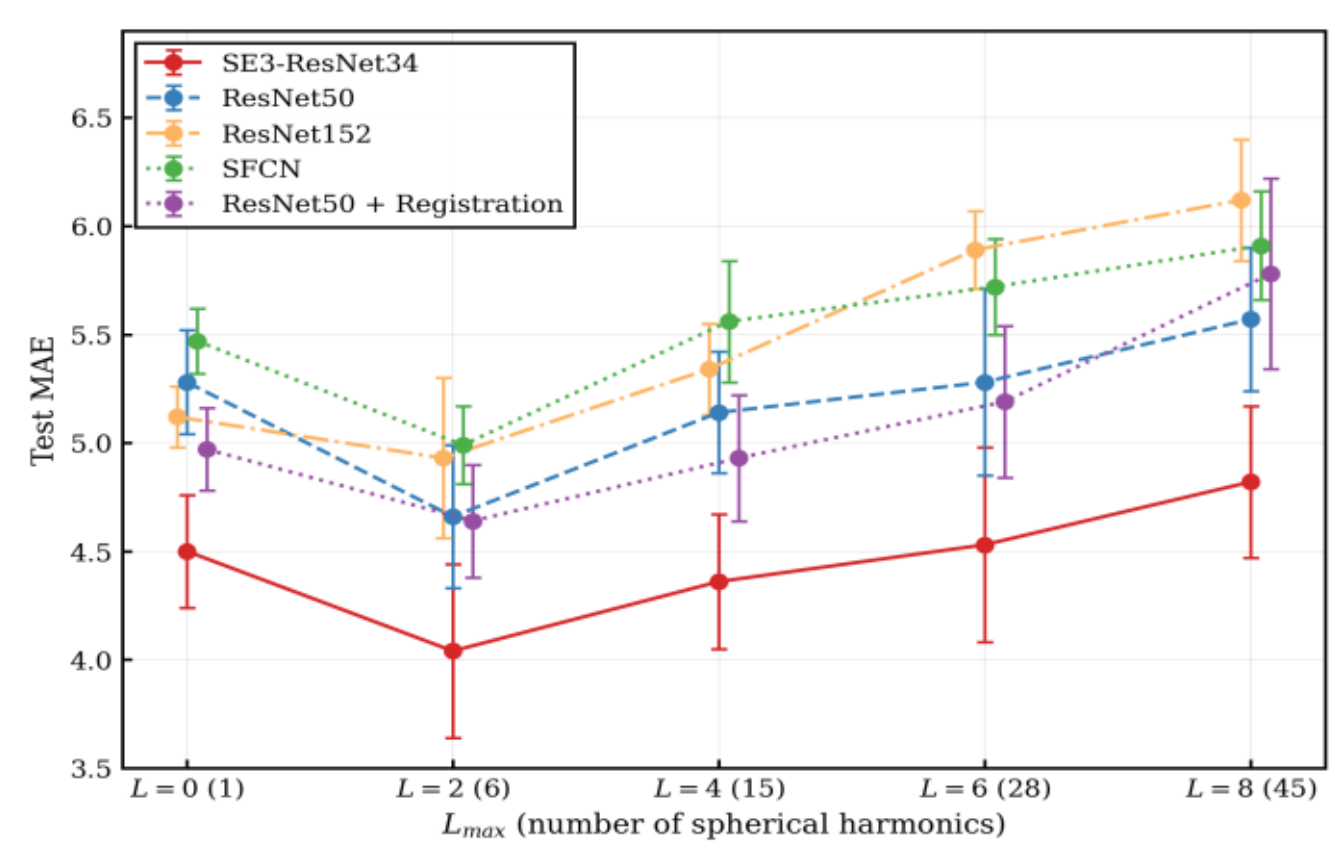

Fig. 3. Test-set MAE across maximum SH orders (Lmax) for SE(3)-ResNet34, ResNet50, registered-input ResNet50, ResNet152, and SFCN. Points are slightly offset to improve visibility. All architectures achieved their lowest MAE at Lmax = 2, where SE(3)-ResNet34 achieved 4.04 ± 0.40 years. Error bars show the SD of test-set MAE across the five cross-validation models.

### 3) *Effect of Network Architecture*

At Lmax = 2, SE(3)-ResNet34 achieved an MAE of 4.04 ± 0.40 years and r = 0.942, outperforming ResNet50 ($p = 0.006$), ResNet152 ($p < 0.001$), and SFCN ($p < 0.001$). SE(3)-ResNet34 achieved the highest accuracy at every maximum SH order.

To test whether spatial normalization could substitute for architectural equivariance, we compared SE(3)-ResNet34 applied to unregistered WM-FODs with ResNet50 applied to WM-FODs nonlinearly registered to a tissue-unbiased FOD template [35] using MRtrix3 mrregister with FOD reorientation (Table 2). Registration did not significantly improve ResNet50 at any Lmax (all $p > 0.05$), and SE(3)-ResNet34 without registration outperformed registered ResNet50 at every evaluated order (all $p < 0.05$).

TABLE 2. WM-FOD BRAIN AGE PREDICTION MAE ACROSS MAXIMUM SH ORDERS AND SELECTED NETWORK CONFIGURATIONS

| Lmax | SE(3)-ResNet34 | ResNet50 | ResNet50 (registered) |
|---|---|---|---|
| 0 | 4.50 ± 0.26 | 5.28 ± 0.24 | 4.97 ± 0.19 |
| 2 | **4.04 ± 0.40** | 4.66 ± 0.33 | 4.64 ± 0.26 |
| 4 | 4.36 ± 0.31 | 5.14 ± 0.28 | 4.93 ± 0.29 |
| 6 | 4.53 ± 0.45 | 5.28 ± 0.43 | 5.19 ± 0.35 |
| 8 | 4.82 ± 0.35 | 5.57 ± 0.33 | 5.78 ± 0.44 |

Values are test-set MAE (years), given as mean ± SD across the five cross-validation models (n = 104). Bold indicates the lowest MAE. Registered: WM-FODs nonlinearly registered to the FOD template.

### 4) *Combining WM-FODs with Equivariant Processing*

The best WM-FOD model (SE(3)-ResNet34, 4.04 years; Table 2) improved on the DTI benchmark (ResNet50 with early fusion, 4.97 years; Table 1) by 0.93 years. To determine whether this gain arose from the input features, the network architecture, or their combination, we compared configurations that differed in only one component. With ResNet50 held fixed, replacing DTI early fusion with WM-FODs (Lmax = 2) did not significantly improve accuracy (4.66 versus 4.97 years; $p = 0.35$). With DTI early fusion held fixed, replacing ResNet50 with SE(3)-ResNet34 also produced no significant improvement (4.92 ± 0.13 versus 4.97 years; $p = 0.73$). Accuracy improved significantly only when WM-FODs were combined with SE(3)-ResNet34.

### 5) *Ensemble Prediction Performance*

Averaging predictions from the five cross-validation models improved performance across configurations. For SE(3)-ResNet34 at Lmax = 2, ensemble averaging reduced MAE from 4.04 to 3.54 years and yielded r = 0.958. The baseline models also benefited from averaging (Fig. 4), while the SE(3)-ResNet34 ensemble retained significantly better performance than all baseline ensembles (all $p < 0.05$).

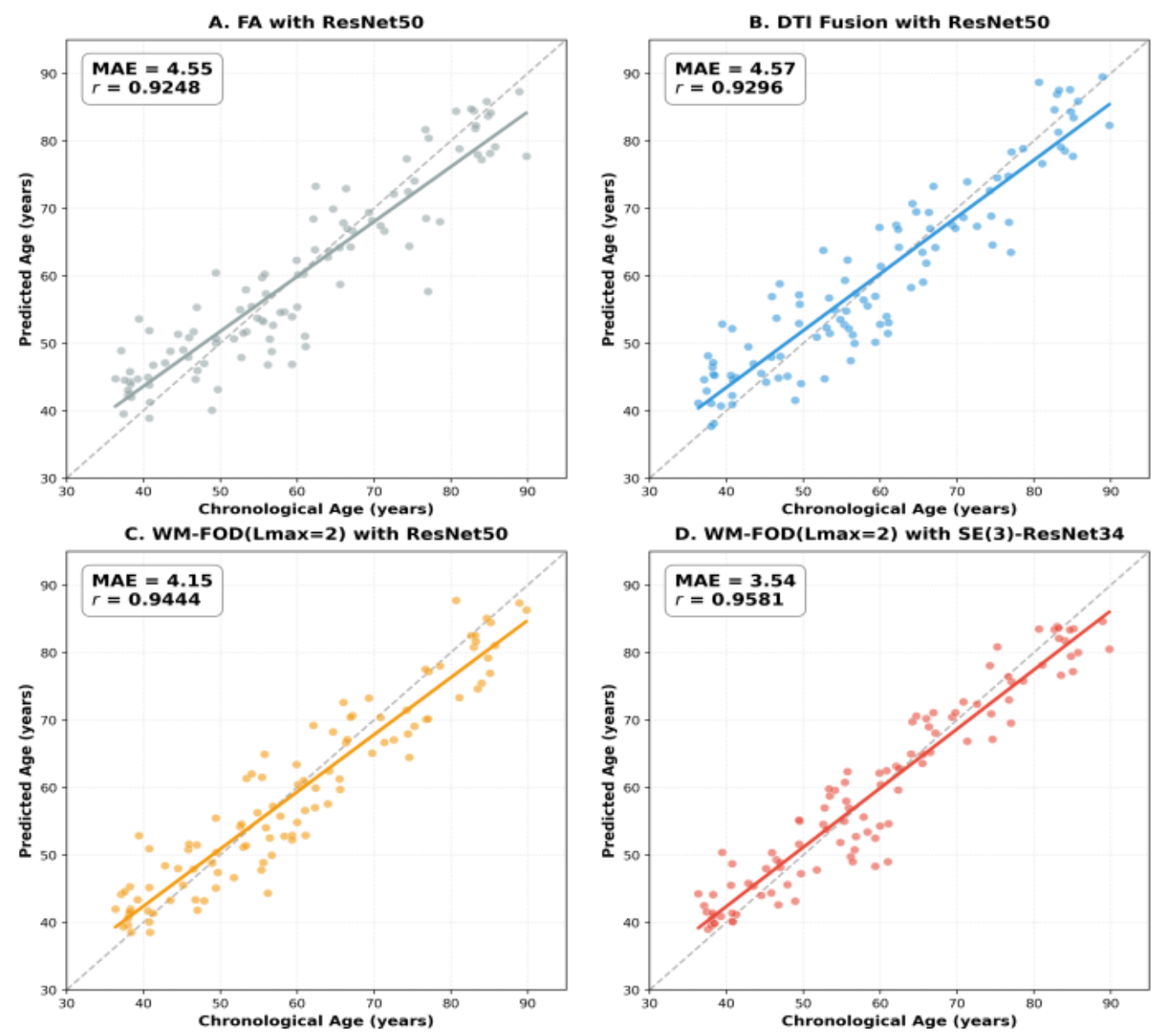

Fig. 4. Ensemble-predicted versus chronological age in the held-out test set (n = 104): (a) DTI FA baseline, (b) DTI early fusion, (c) ResNet50 with WM-FODs, and (d) SE(3)-ResNet34 with WM-FODs. Each prediction is the average of five cross-validation model predictions. The dashed gray line denotes identity (y = x), and solid colored lines show linear fits. The SE(3)-ResNet34 ensemble achieved the lowest prediction error.

## B. *Architectural Evaluation*

### 1) *Prediction Stability Under Rotation*

Figure 5 illustrates prediction stability for a test participant whose unrotated absolute error was 0.5 years for both SE(3)-ResNet34 and ResNet50. Across rotations of up to ±15° in the axial, coronal, and sagittal planes, SE(3)-ResNet34 predictions had an SD of 0.28 years. ResNet50 predictions were more variable (SD = 2.41 years), with absolute errors exceeding 8 years at −15° coronal rotation. Thus, the prediction SD was 8.6 times greater for ResNet50 in this example.

The analysis of all test participants (n = 104) showed the same pattern. SE(3)-ResNet34 had a mean rotation-induced prediction SD of 0.90 ± 0.40 years, significantly lower than ResNet50 (2.99 ± 0.89 years; $p < 0.001$) and rotation-augmented ResNet50 (2.04 ± 0.45 years; $p < 0.001$). Augmentation improved stability relative to unaugmented ResNet50 ($p < 0.001$), but did not significantly improve prediction accuracy (MAE = 4.67 ± 0.28 versus 4.66 ± 0.33 years; $p = 0.83$).

### 2) *Computational Cost*

SE(3)-ResNet34 required 406 seconds per training epoch, approximately 1.6 times the 258 seconds required by ResNet50, compared with 265 seconds for ResNet152 and 231 seconds for SFCN. With 2.95 million parameters in the configuration of Fig. 2, SE(3)-ResNet34 was approximately 16 and 40 times smaller than ResNet50 (46.27 million) and ResNet152 (117.47 million), respectively, and comparable in size to SFCN (2.95 million), while achieving the best prediction performance. The equivariant architecture therefore achieved its accuracy with far fewer parameters than the ResNet baselines, at a higher computational cost per epoch.

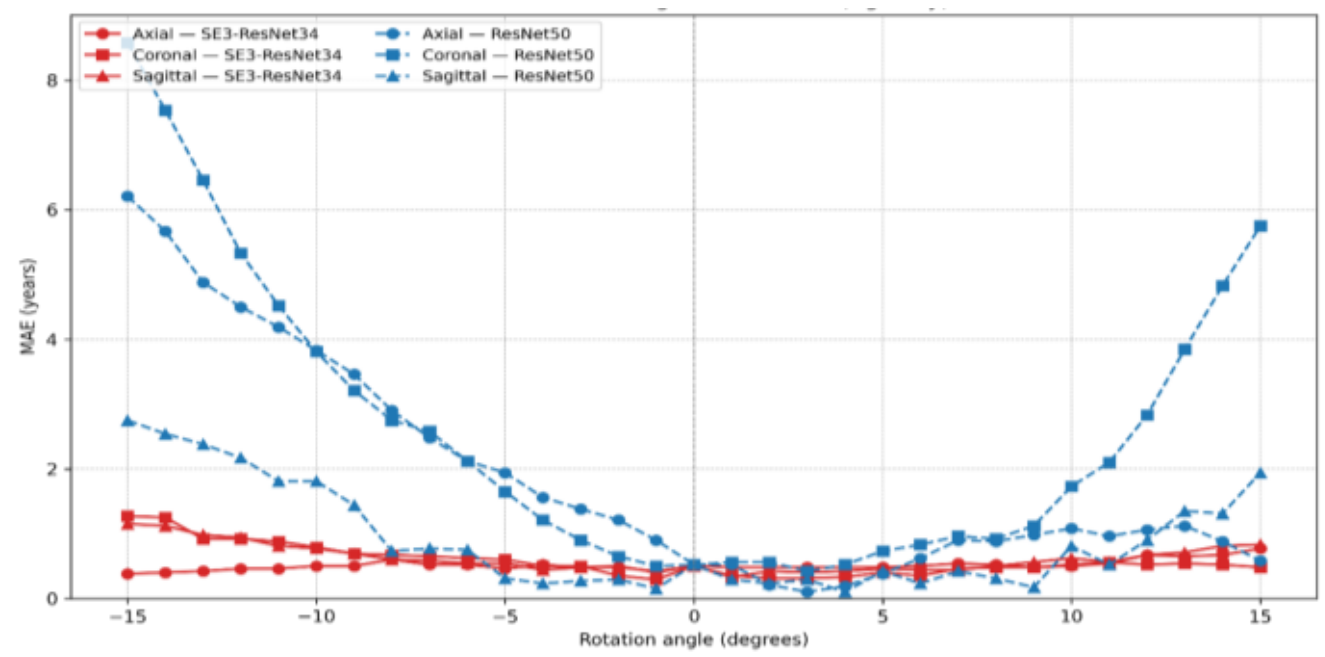

Fig. 5. Prediction stability under input rotation. Absolute age prediction error is shown for rotations from −15° to +15° in the axial (circles), coronal (squares), and sagittal (triangles) planes for an illustrative test participant. Both models had an unrotated absolute error of 0.5 years. SE(3)-ResNet34 (red, solid) remained comparatively stable, whereas ResNet50 (blue, dashed) showed errors exceeding 8 years at −15° coronal rotation.

## C. Tract-Level Importance

Tract-level occlusion quantified model sensitivity using two complementary measures: the mean absolute change in prediction error after occlusion (|ΔMAE|) and its volume-normalized counterpart. The highest values (Fig. 6A) occurred in corpus callosum subregions CC_2 (genu; |ΔMAE| = 3.99 years) and CC_6 (isthmus; |ΔMAE| = 3.92 years), and the middle cerebellar peduncle (MCP; |ΔMAE| = 3.81 years). Several projection pathways also ranked highly, including the bilateral prefrontal thalamic radiations (T_PREF), fronto-pontine tracts (FPT), and anterior thalamic radiation (ATR).

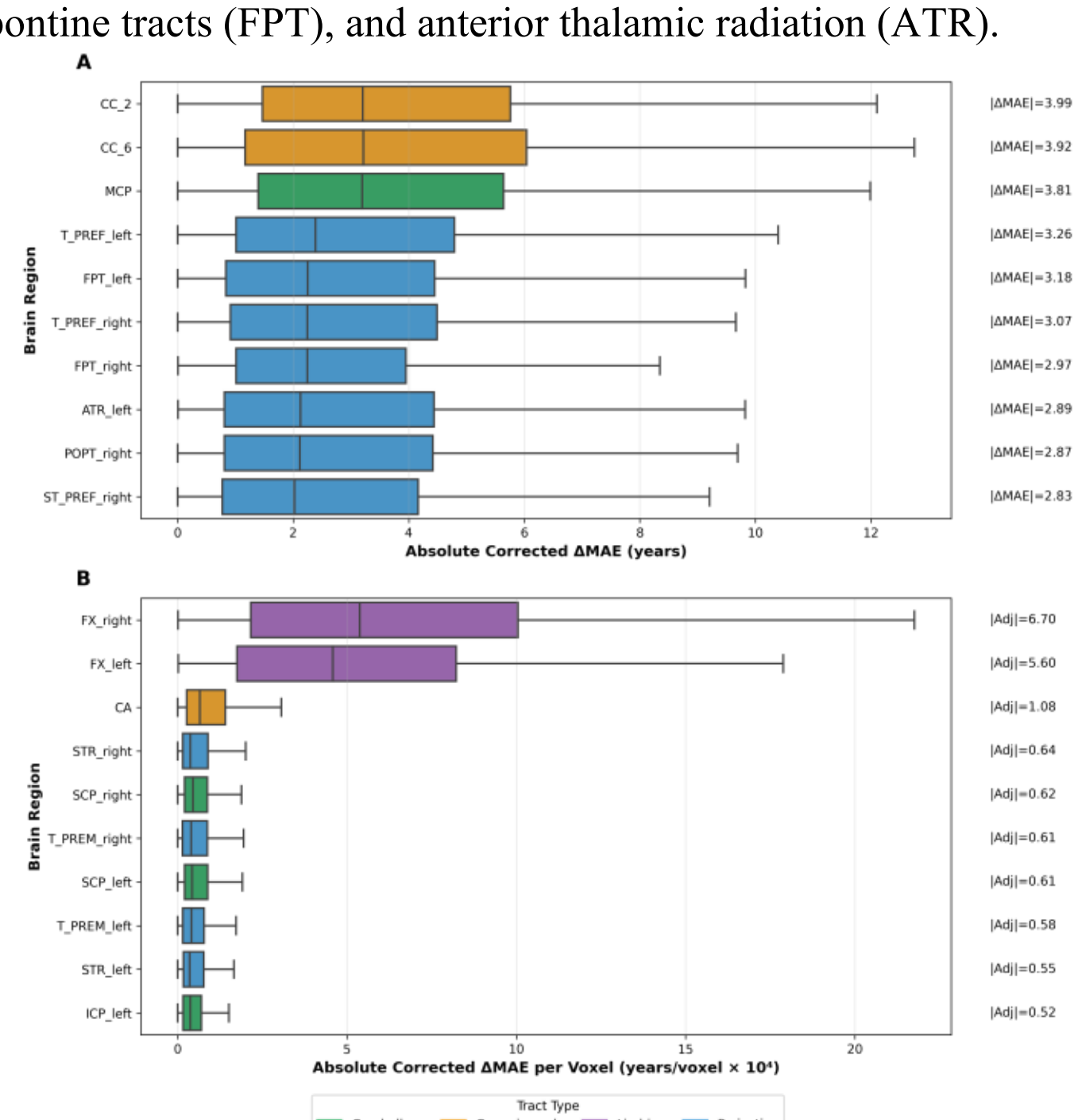

Fig. 6. Tract-level model sensitivity. (A) Absolute change in age-bias-corrected prediction error after tract occlusion. Box plots show participant-level values (n = 692); numbers at right give the mean (|ΔMAE|, years). CC_2 (genu), CC_6 (isthmus), and the MCP showed the largest effects. (B) Per-voxel importance, calculated by normalizing |ΔMAE| by tract volume (×$10^{-4}$ years/voxel); the bilateral fornix ranked highest. The 10 highest-ranked tracts are shown for each measure and colored by anatomical category. Box plots indicate the median and interquartile range (IQR), with whiskers extending to 1.5 × IQR; outliers are not shown.

Normalizing |ΔMAE| by tract volume reduced the influence of tract size on the importance ranking (Fig. 6B). The bilateral fornix had the highest per-voxel importance (FX_right: 6.70 × $10^{-4}$ years/voxel; FX_left: 5.60 × $10^{-4}$ years/voxel), followed by the anterior commissure (CA: 1.08 × $10^{-4}$ years/voxel). All remaining tracts had values below 0.65 × $10^{-4}$ years/voxel.

## D. Age-Stratified Tract Importance

We compared tract importance across five age bins to characterize age-related variation in model sensitivity (Fig. 7). Tract importance maps and streamlines were displayed on the tissue unbiased brain template [35]. For each tract, we identified the age bin with the highest importance; for each bin, we report the tracts with the largest peak values. These comparisons are descriptive.

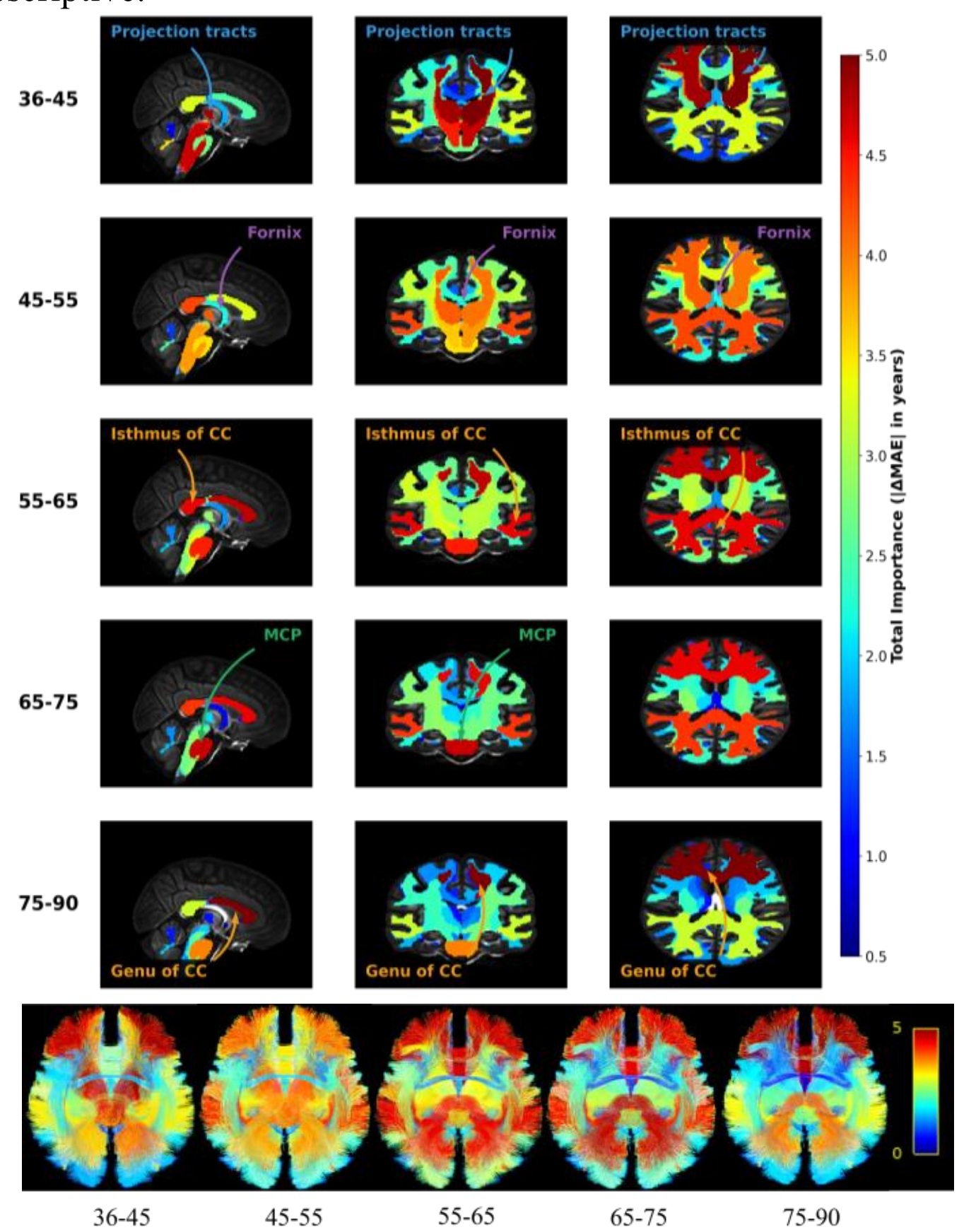

Fig. 7. Tract importance across five age bins spanning 36–90 years. Selected importance peaks occur in frontal projection pathways at 36–44 years, the fornix at 45–54 years, the callosal isthmus at 55–64 years, the MCP at 65–74 years, and the callosal genu at 75–90 years. Bottom row: streamlines colored by importance. All panels use the same color scale. Age-group differences describe model sensitivity rather than longitudinal changes within individuals.

Early midlife (36–44 years): the largest peaks were in frontal projection pathways, namely, the left prefrontal thalamic radiation (|ΔMAE| = 5.08 years), left fronto-pontine tract (|ΔMAE| = 4.72 years), and left anterior thalamic radiation (|ΔMAE| = 4.65 years); their right-hemisphere counterparts also peaked in this bin (4.84, 4.47, and 4.36 years, respectively). Midlife (45–54 years): the fornix reached its highest importance (left |ΔMAE| = 2.14 years; right 1.91 years). Late midlife (55–64 years): the isthmus of the corpus callosum (CC_6) reached its highest importance (|ΔMAE| = 4.57 years). Early aging (65–74 years): the middle cerebellar peduncle reached its highest

importance (|ΔMAE| = 4.70 years). Late aging (75–90 years): the genu of the corpus callosum (CC_2) reached its highest importance (|ΔMAE| = 5.03 years).

## V. Discussion

This study examined how diffusion feature representation and geometric network architecture contribute to dMRI-based brain age prediction. Three findings emerged. First, the combination of WM-FODs and SE(3)-equivariant processing achieved the best prediction performance; changing either component alone did not significantly improve accuracy in comparisons that varied one component at a time. Second, accuracy peaked at Lmax = 2, and the equivariant model was more stable under input rotations than conventional CNNs. Third, tract-level occlusion identified age-dependent patterns of model sensitivity, with particularly high volume-normalized importance in the bilateral fornix.

### A. WM-FOD Representations for Brain Age Prediction

FA is the most widely used DTI scalar input for age prediction and yielded the lowest error among the individual scalar metrics in our experiments, while combining scalar metrics further improved prediction accuracy, consistent with the value of complementary diffusion information [5], [16]. Comparisons that varied one component at a time also suggest that the benefit of WM-FODs depends on how the network processes them. Replacing DTI fusion with WM-FODs within ResNet50 did not significantly improve accuracy, and applying SE(3)-ResNet34 to DTI fusion likewise produced no significant gain. The strongest performance was obtained when the orientation-dependent WM-FOD representation was processed with an architecture designed to respect its geometry, i.e., SE(3)-ResNet34.

Several properties may contribute to this result. MSMT-CSD relaxes the single-Gaussian assumption of DTI and separates white matter, gray matter, and CSF signal components [20]. WM-FOD SH coefficients also retain orientation information that is absent from rotationally invariant DTI scalar maps. Previous studies have shown that FOD-derived measures are sensitive to age-related white matter differences [36], [37].

We observed a non-monotonic relationship between SH order and prediction performance, with accuracy peaking at Lmax = 2 and declining at higher orders. One possible explanation is that increasing Lmax from 2 to 8 raises input dimensionality from 6 to 45 channels, which may make generalization more difficult with 588 training participants. At the same time, age-relevant information may be concentrated at low orders: healthy aging is characterized by widespread reductions in fiber density [36] and anisotropy [38], which at the voxel level may be largely expressed in the $\ell = 0$ (total fiber density) and $\ell = 2$ terms. Moreover, although Lmax = 2 does not resolve crossing-fiber populations as distinct peaks, its coefficients, being a linear projection of the FOD, still change with the relative contribution and dispersion of fiber populations within a voxel, and may therefore capture age-related changes without explicit peak resolution.

The SE(3)-ResNet34 ensemble achieved an MAE of 3.54 years and r = 0.958, comparing favorably with published unimodal dMRI-based brain age studies in healthy aging cohorts (Fig. 8). Note however that such comparisons provide general context rather than a controlled ranking, because performance depends on cohort size, age range, acquisition, preprocessing, and evaluation procedures.

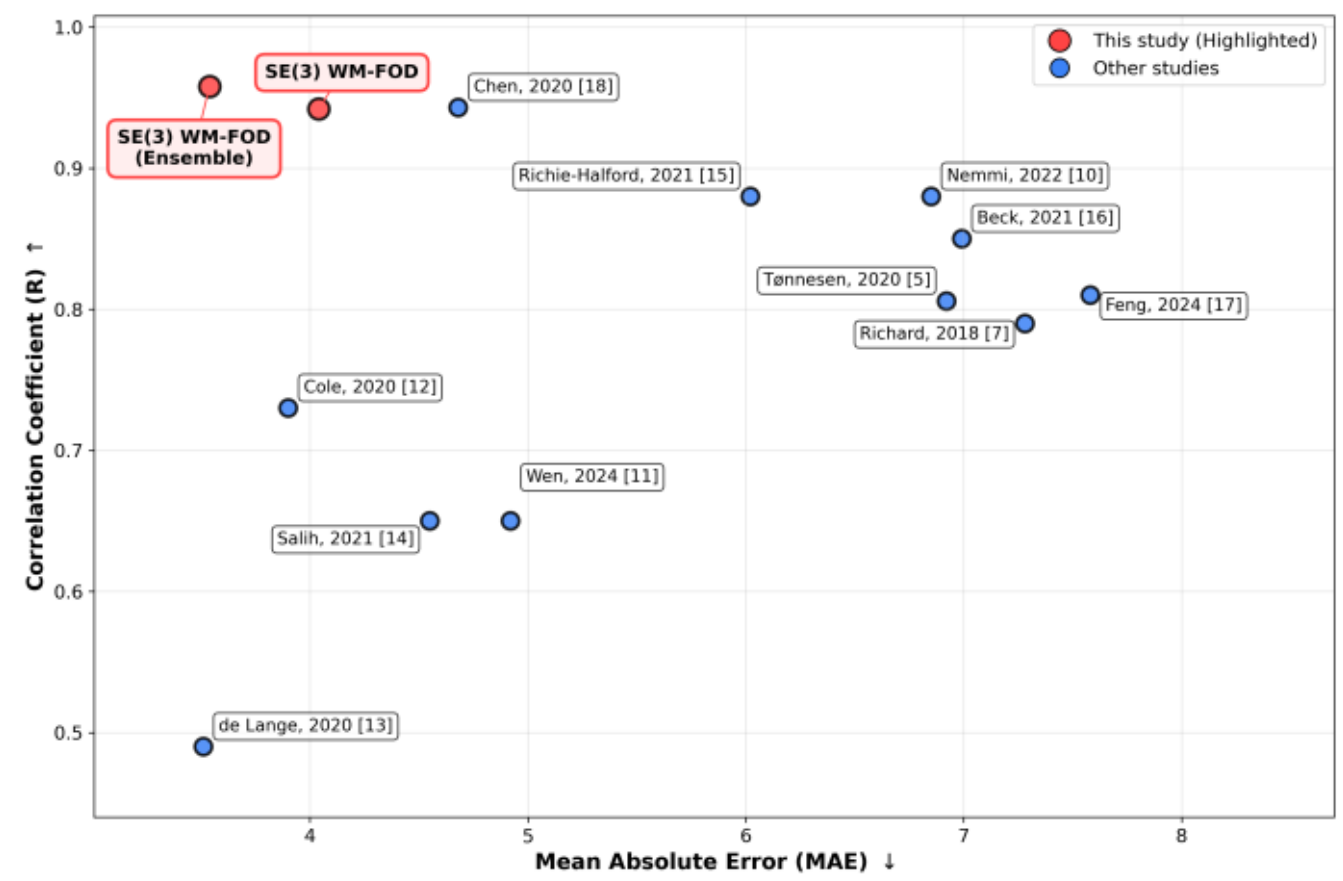

Fig. 8. Comparison with unimodal dMRI-based brain age studies in healthy aging cohorts that include adults older than 65 years. Red markers indicate SE(3)-ResNet34 performance with and without ensemble averaging. Differences in cohorts and evaluation procedures should be considered when interpreting these comparisons.

### B. SE(3)-Equivariant Processing of WM-FODs

SE(3)-equivariant processing improved prediction from WM-FODs relative to the conventional CNNs. Previous equivariant dMRI methods include the joint SE(3) × SO(3) convolutions of Müller et al. for raw q-space signals [22] and the E(3) × SO(3) hemispherical convolutions of Elaldi et al. [23]. These approaches jointly model the spatial and angular domains. Our network instead processes SH coefficients estimated in advance using MSMT-CSD. This compact angular representation maps directly to the irreducible representations used by SE(3)-steerable kernels, enabling equivariant processing without an additional SO(3) angular convolution. To our knowledge, this is the first application of a volumetric SE(3)-equivariant CNN directly to whole-brain WM-FOD SH coefficient images for brain age prediction.

The gains in accuracy and rotational stability are consistent with better alignment between the network architecture and WM-FOD geometry. Standard CNNs mix SH coefficients without enforcing how each SH order transforms under rotation, whereas SE(3)-equivariant networks preserve these transformations. This may also improve sample efficiency: similar fiber configurations occur at different orientations across tracts and participants, and equivariant kernels share weights across orientations rather than learning each separately [9]. Consistent with this, SE(3)-ResNet34 was more stable under rotation than ResNet50; rotation augmentation narrowed but did not close this gap and did not significantly improve accuracy.

Registration did not substitute for equivariance: SE(3)-ResNet34 applied without registration outperformed ResNet50 applied to registered WM-FODs at every evaluated SH order.

This benefit came with a computational trade-off. SE(3)-ResNet34 required approximately 1.6 times the training time per epoch of ResNet50, but had only 2.95 million parameters, approximately 1/16 of the ResNet50 parameter count. SFCN provided a parameter-matched comparison: with a similar number of parameters, it performed significantly worse, so the gain cannot be attributed to model size.

### C. Age-Dependent Model Sensitivity to White Matter Regions

Tract-level occlusion identified age-dependent patterns of model sensitivity that can be considered alongside previous findings on white matter aging.

In early midlife (36–44 years), the largest peaks occurred in frontal projection pathways: the prefrontal thalamic radiation, fronto-pontine tract, and anterior thalamic radiation. Diffusion-derived measures of white matter tracts have been reported to peak between approximately 20 and 40 years [38], followed by age-related reductions in white matter integrity [39]. This bin may therefore capture the onset of their decline, although occlusion does not identify which microstructural properties drive their importance.

The bilateral fornix had the highest per-voxel importance and peaked in midlife (45–54 years). Previous work has identified pronounced age-related reductions in fornix FA [40]. As a principal hippocampal output pathway, the fornix has also been linked to memory performance [41] and progression from mild cognitive impairment to dementia [42], and its degeneration has been suggested to precede hippocampal dysfunction [43]. Its small size and proximity to the ventricles, however, make it susceptible to CSF partial-volume effects that increase with age, which MSMT-CSD reduces but may not fully remove.

The MCP reached its highest importance at 65–74 years (Fig. 7). Age-related differences in the cerebellum have been linked to motor and cognitive decline [44], although evidence specific to the MCP is limited.

The callosal isthmus and genu reached their highest importance at 55–64 and 75–90 years, respectively. Age-related white matter decline has been reported to be more pronounced in anterior regions [45], [46], which include the prefrontal territory connected by the genu. However, the timing of peak predictive importance does not directly indicate the timing or rate of tissue degeneration.

Across age groups, the most prominent sensitivities involved specific projection, limbic, commissural, and cerebellar bundles. Their correspondence with previously reported age-related differences supports the plausibility of the model's feature use. These importance patterns should nevertheless be interpreted as properties of the trained model: they do not establish causal mechanisms of aging or directly measure cognitive function.

### D. Limitations and Future Work

Several limitations warrant consideration. First, evaluation was restricted to one healthy aging cohort; generalization across scanners, acquisition protocols, and clinical populations remains to be established. Second, voxel-wise tract occlusion also removes signals from any crossing fiber populations within the mask, so importance reflects sensitivity to tract-defined regions rather than isolated bundles. This confounding effect increases for larger tract volumes, as more sections of unrelated tracts will also be occluded. Third, normalization by tract volume reduces the influence of tract size but does not account for tract topology or overlap. External validation and more fiber-specific perturbation methods are therefore important directions for future work.

## VI. Conclusion

We presented an SE(3)-equivariant framework for brain age prediction from WM-FODs. Processing WM-FODs with an SE(3)-equivariant network yielded the best performance among the evaluated configurations, with an ensemble MAE of 3.54 years and r = 0.958. The model also showed greater stability under input rotations and used approximately 1/16 of the parameters of ResNet50. Comparisons that varied one component at a time showed that the only significant gain was obtained when WM-FOD inputs were combined with equivariant processing, rather than with either change alone. Tract-level occlusion further identified age-dependent patterns of model sensitivity to white matter regions. These findings support SE(3)-equivariant networks, which preserve how WM-FOD SH coefficients transform under rotation, as a promising approach to accurate and interpretable dMRI-based brain age prediction.

## ACKNOWLEDGMENT

Research reported in this publication was supported by the National Institute On Aging of the National Institutes of Health under Award Number U01AG052564 and by funds provided by the McDonnell Center for Systems Neuroscience at Washington University in St. Louis. The HCP-Aging 2.0 Release data used in this report came from DOI: 10.15154/1520707. J. Lv is supported by Brain and Mind Centre Research Development Grant, USYD-Fudan Brain and Intelligence Science Alliance Flagship Research Program, Moyira Elizabeth Vine Fund for Research into Schizophrenia Program. JL, FC and SN are supported by an ARC Discovery Project (DP240102161).